\documentclass[10pt]{article} 
\usepackage{arxiv}
\usepackage{spconf,amsmath,graphicx,booktabs,float, placeins}
\usepackage{url}

\title{Learning Single-Image Super-Resolution in the JPEG Compressed Domain}
\name{
Sruthi Srinivasan, 
Elham Shakibapour, 
Rajy Rawther,
Mehdi Saeedi
}
\address{
Advanced Micro Devices, Inc.
}

\begin{document}
%
\maketitle
\begin{abstract}
Deep learning models have grown increasingly complex, with input data sizes scaling accordingly. Despite substantial advances in specialized deep learning hardware, data loading continues to be a major bottleneck that limits training and inference speed. To address this challenge, we propose training models directly on encoded JPEG features, reducing the computational overhead associated with full JPEG decoding and significantly improving data loading efficiency. While prior works have focused on recognition tasks, we investigate the effectiveness of this approach for the restoration task of single-image super-resolution (SISR). We present a lightweight super-resolution pipeline that operates on JPEG discrete cosine transform (DCT) coefficients in the frequency domain. Our pipeline achieves a 2.6x speedup in data loading and a 2.5x speedup in training, while preserving visual quality comparable to standard SISR approaches.
\end{abstract}


\section{Introduction}
\label{sec:intro}
The efficiency of deep learning pipelines hinges on how quickly input data is processed. In image-based tasks, neural networks rely on RGB pixels, matching common display formats. However, images are often stored in compressed formats such as JPEG, requiring reading and preprocessing, also known as data loading \cite{zolnouri2020importance}. Due to the increased complexity and size of image datasets, inefficiencies in the data-loading phase are particularly important \cite{yang2019accelerating}. In the regular data-loading pipeline, CPU reads input images from disk, decodes them from formats like JPEG, and applies preprocessing before transferring data to the GPU to be used for training or inference. This process can lead to under-utilization of GPU resources, as the CPU struggles to keep pace with the training speed, contributing to data loading inefficiencies that can account for up to 40\% of the epoch time \cite{zolnouri2020importance}.

\begin{figure}
    \centering
    \includegraphics [width=1\linewidth]{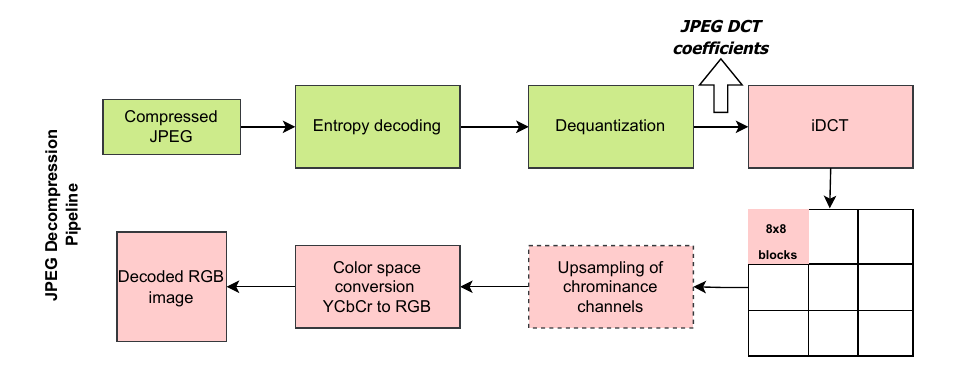}
    \caption{JPEG decompression pipeline showing the use of DCT coefficients without full decoding to RGB image.}
    \label{fig:dataloading}
\end{figure}

Restoring RGB images from JPEG-compressed data involves reversing the JPEG compression process. A core component of this is frequency-domain transformation using DCT as specified in the JPEG standard (ISO/IEC 10918-1:1993). This lossy compression format is widely adopted in digital imaging. By leveraging the frequency-domain DCT coefficients directly, we can bypass the computationally expensive decoding step required to convert JPEG data into RGB format (Fig. \ref{fig:dataloading}). This approach accelerates the data loading process and optimizes the entire end-to-end training pipeline. Directly using the frequency-domain coefficients also allows deep learning models to operate on a compressed representation that is 1/8th the size of the original RGB image in both height and width, further reducing memory and computational requirements.

Although frequency-domain deep learning has proven effective for recognition tasks \cite{gueguen2018faster,ehrlich2019deep}, its application to pixel-level restoration tasks like super-resolution remains underexplored. Extending frequency-domain learning to single-image super-resolution (SISR) requires reconstructing high-resolution (HR) images from low-resolution (LR) inputs, which demands finer precision for texture and detail reconstruction compared to recognition tasks. Deep learning-based SISR approaches include CNN-based \cite{zhang2021edge,lim2017enhanced}, transformer-based \cite{liang2021swinir,lu2022transformer}, and GAN-based methods \cite{ledig2017photo}. Among them, CNN-based methods offer a balance in speed and quality \cite{chen2022mobile}, making them suitable for resource-constrained environments -- a use case targeted in our study.  

In this paper, we propose a lightweight CNN-based SISR pipeline that operates directly on JPEG discrete cosine transform (DCT) coefficients, eliminating the need for full decoding to RGB. This design significantly reduces data loading overhead and accelerates end-to-end training while maintaining competitive super-resolution quality. In summary, our contributions are two-fold \textbf{(1)} A lightweight SISR pipeline designed for JPEG-DCT frequency-domain inputs, enabling fast data loading, accelerated end-to-end training, and efficient inference. \textbf{(2)} An evaluation of quality and speed trade-offs in frequency-domain SISR using JPEG-DCT coefficients.

\section{Related Work}
\label{sec:relatedwork}
The increasing size of datasets and complexity of deep learning models have motivated research into training acceleration. Model-level optimizations such as depthwise separable convolutions \cite{howard2017mobilenets}, factorized convolutions \cite{iandola2016squeezenet}, quantization \cite{jacob2018quantization}, and low-rank approximations \cite{denton2014exploiting} have demonstrated substantial reductions in model size and computational requirements. In parallel, bottlenecks in the data-loading pipeline -- especially those arising from CPU-based decoding and augmentation have been addressed by GPU-accelerated libraries such as NVIDIA DALI \cite{nvidia_dali} and AMD’s rocAL \cite{rocAL}. 

Despite these advances, JPEG-to-RGB decoding remains a substantial computational overhead. To address this limitation, several studies have explored training directly on JPEG DCT coefficients in the frequency domain. Works such as \cite{gueguen2018faster, xu2020learning, lo2019exploring, deguerre2019fast} have successfully adapted CNN architectures to process JPEG DCT coefficients, achieving consistent speedups for recognition tasks including classification, object detection, and segmentation. However, their applicability to pixel-wise restoration tasks like super-resolution remains largely unexplored.

In this paper, we extend frequency-domain processing to SISR, which reconstructs HR images from LR inputs. Our approach processes JPEG DCT coefficients directly, eliminating decoding overhead and accelerating both data loading and training while maintaining quality -- particularly beneficial for resource-constrained environments.

\section{Proposed Method}
\label{sec:proposedmethod}
JPEG compression \cite{wallace1991jpeg} encodes images by first converting them to the YCbCr color space, dividing them into 8×8 blocks and applying DCT to convert spatial-domain pixels into frequency-domain coefficients. These coefficients are then quantized to discard perceptually less important and high-frequency components. 
The resulting DCT coefficients are organized by frequency content: the DC coefficient captures the average intensity of each block, while the AC coefficients represent finer spatial variations. JPEG stores these coefficients separately for the luminance (Y) and chrominance (Cb, Cr) components. The Y channel has shape (1, H/8, W/8, 8, 8), while the subsampled Cb and Cr channels have shape (2, H/16, W/16, 8, 8), with each 8×8 block containing 64 coefficients. JPEG compression provides major storage efficiency through this frequency-domain representation.

\begin{figure*}[tb]
    \centering
    \includegraphics [width=1\textwidth]{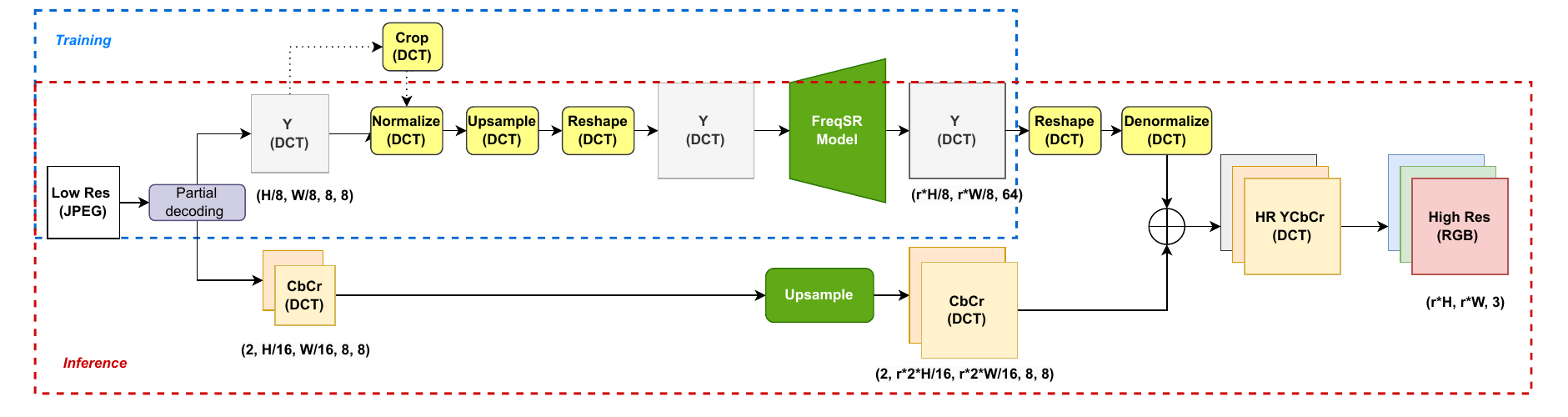}
    \caption{The proposed end-to-end SISR pipeline operates on JPEG DCT coefficients as frequency domain input.}
    \label{fig:architecture}
\end{figure*}

The proposed super-resolution pipeline operates directly on JPEG DCT coefficients in the frequency domain. This approach eliminates the computational overhead of JPEG decoding and RGB reconstruction, delivering both storage efficiency and faster data loading during training and inference. Fig.~\ref{fig:architecture} presents the complete pipeline architecture. The following sections detail the four key components: input processing, preprocessing operations, model architecture, and inference procedures.

\subsection{Input}
Training targets the Y channel of JPEG DCT coefficients exclusively, since this luminance component encodes the majority of perceptually relevant information. Human vision is more sensitive to variations in luminance than chrominance, making the Y channel crucial for tasks like super-resolution. Additionally, chrominance channels are upsampled separately during post-processing for inference, allowing enhanced efficiency in training. This Y-channel-focused strategy aligns with common practices in image restoration tasks \cite{liang2021swinir}, where the luminance channel is often prioritized due to its predominant role in visual perception.

\subsection{Preprocessing}
Given an LR JPEG image, Y DCT coefficients are extracted and processed through the following operations:

\begin{itemize}
\item{\textbf{Cropping}}: A centered patch of size \(S \times S\) is cropped from the low-resolution Y DCT tensor to ensure uniform processing. 

\item {\textbf{Normalizing}}: Following cropping, the LR Y DCT coefficients are normalized to the range \( [-1, 1] \) to facilitate stable training. This normalization applies the transformation defined in Eq. \ref{eq:normalization}.

\begin{equation}
val_{\text{min}} + \left( \frac{x - orig_{\text{min}}}{orig_{\text{max}} - orig_{\text{min}}} \right) \cdot (val_{\text{max}} - val_{\text{min}})
\label{eq:normalization}
\end{equation}
where \( x \) is the original DCT coefficient. The original DCT coefficient range \([-1024, 1016]\), as in \cite{park2023rgb}, is mapped to the normalized range \([-1, 1]\) using 
\(\text{orig}_{\text{min}} = -1024\), \(\text{orig}_{\text{max}} = 1016\), \(\text{val}_{\text{min}} = -1\), and \(\text{val}_{\text{max}} = 1\).

\item{\textbf{Upsampling}}: The normalized Y DCT coefficients are upsampled by the target scale factor to increase both spatial dimensions. This upsampling follows the methodology in \cite{park2023rgb}, where DCT coefficients are padded and scaled to enlarge block size, then decomposed using conversion matrices and rearranged into the final upsampled configuration. Unlike conventional approaches that perform upsampling during training, the preprocessing-based upsampling strategy from \cite{cai2021freqnet} is adopted to better preserve local neighborhood features and retain fine-grained details.

\item {\textbf{Reshaping}}: The Y DCT coefficients are structured as a 5D tensor (1, H/8, W/8, 8, 8), corresponding to batch size, block height, block width, and the 8×8 DCT block dimensions. For model compatibility, the last two dimensions are flattened to create a 4D tensor of shape (1, H/8, W/8, 64), preserving the spatial layout while converting each 8×8 block into a 64-element feature vector.
\end{itemize}

\begin{figure}[H]
    \centering
    \includegraphics[width=0.7\linewidth]{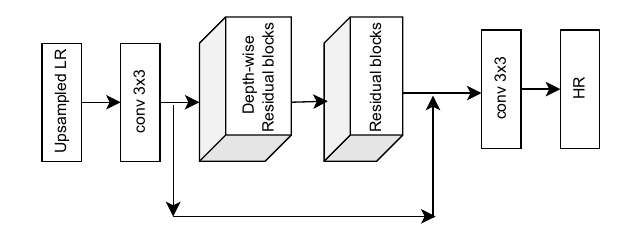}
    \caption{Overview of the FreqSR model architecture. The architecture uses a convolutional layer for feature extraction, depth-wise residual blocks for independent channel processing, and standard residual blocks for cross-channel refinement. Operating on reduced spatial dimensions, the model achieves faster training.}
    \label{fig:model}
\end{figure}

\subsection{Model Architecture}
The end-to-end SISR pipeline employs a frequency-domain super-resolution (FreqSR) model, illustrated in Fig. \ref{fig:model}. The architecture inspired from EDSR \cite{lim2017enhanced}, begins with a convolutional layer that extracts features from the upsampled LR input. Depth-wise residual blocks \cite{cai2021freqnet} process each channel independently, preserving the integrity of individual frequency components since each of the 64 channels represents a distinct frequency. This design minimizes unnecessary mixing across frequencies. Standard residual blocks then refine these features and model cross-channel frequency relationships to enhance reconstruction accuracy. A final convolutional layer produces the output. By operating on inputs with spatial dimensions reduced to 1/8th the height and width of those used in RGB methods, the model achieves faster training.

\subsection{Inference}
During inference, the preprocessed DCT Y channel (described in Section 3.2) is fed into the trained model. The model output undergoes several post-processing steps: (1) reshaping the flattened dimensions back to the original 8×8 block structure, (2) denormalizing the coefficients to restore their original range, (3) upsampling the Cb and Cr channels to match the Y channel scale, and (4) merging all channels and converting back to RGB format for final output.

\section{Experiments}
\label{sec:experiments}
We evaluate our proposed SISR pipeline through different experiments. For clarity, we define three key components: (1) FreqSR model -- the neural network architecture trained directly on JPEG DCT coefficients in the frequency domain; (2) FreqSR training pipeline -- the complete end-to-end training workflow encompassing JPEG DCT data loading and model training; and (3) FreqSR inference pipeline -- the full inference process including post-processing and conversion from JPEG DCT coefficients back to RGB outputs.

\subsection{Dataset}
We use the DIV2K dataset \cite{timofte2017ntire} to train our model. DIV2K is a HR (2K) image dataset specifically designed for image restoration tasks, comprising 800 training images and 100 validation images. To evaluate the SISR pipeline performance, experiments are conducted on the Set5 dataset \cite{bevilacqua2012low}, a widely-used benchmark for single-image super-resolution. Set5 consists of five high-quality images commonly used to assess super-resolution algorithm effectiveness. For consistency across training, validation, and testing, all images in both datasets are transcoded to JPEG format with a quality factor of 100.

\subsection{Implementation Details}
All experiments are conducted using PyTorch. DCT coefficient extraction utilizes a modified version of libjpeg \cite{libjpeg} and TorchJPEG \cite{ehrlich2020quantization}, as provided by \cite{park2023rgb}. Frequency-domain operations -- including cropping, normalization, and upsampling follow the implementation from \cite{park2023rgb}.
The experimental setup uses 32×32 LR block patches in the DCT domain, corresponding to 256×256 LR pixel patches in the RGB domain. HR images are cropped into matching patch sizes with a scale factor of 2 across all experiments. The FreqSR model employs four depth-wise residual blocks and four standard residual blocks. Training uses the Adam optimizer with L1 loss, 
and runs for 100 epochs with batch size 1. All experiments are performed on an NVIDIA RTX 4090 GPU.

\subsection{Data Loading and Training}
Table \ref{tab:training} compares the frequency-domain SISR pipeline using the FreqSR model against spatial-domain SISR baseline approaches (EDSR RGB and EDSR Y) in terms of model complexity, processing speed, and training efficiency. EDSR RGB represents the SISR baseline using all three RGB channels during training, while EDSR Y converts images to YCbCr color space and trains only on the Y channel in the spatial domain. The evaluation encompasses model parameters, decode latency, training data loading speed, and overall training pipeline speed. Both data loading and overall training speeds are averaged over 100 epochs and measured in frames per second (FPS) for consistent performance assessment.

Decoding JPEG DCT coefficients is 65\% faster than decoding images into RGB format, significantly reducing preprocessing overhead. The FreqSR model is also more compact, containing 45\% fewer parameters than EDSR RGB. Depth-wise blocks further improve computational efficiency by minimizing redundant computations. In terms of training performance, the frequency-based SISR pipeline using FreqSR model achieves a 2.6× speedup in data loading compared to EDSR RGB and a 2.48× speedup over EDSR Y. The end-to-end training pipeline for FreqSR model is 2.5× faster than both spatial-domain SISR baselines, highlighting the advantages of frequency-domain processing. This improvement stems from reduced decoding latency, which streamlines data loading and accelerates the complete training pipeline. The experimental results substantiate the practical advantages of frequency-domain training with JPEG DCT representations.

\begin{table*}[tb]
  \centering
  \footnotesize
  \begin{tabular}{lcccc}
    \toprule
    Method & \# of Parameters & Decode latency per image (ms) & Train data loading (FPS) & Train pipeline (FPS) \\
    \midrule
    EDSR RGB & 779,011 & 2.07 & 15.34 & 16.66 \\
    EDSR Y & 776,705 & 2.07 & 16.20 & 16.79 \\
    \textbf{FreqSR (ours)} & 427,776 & 0.72 & 40.27 & 42.64 \\
    \bottomrule
  \end{tabular}
  \caption{Training performance comparison between our proposed SISR pipeline based on FreqSR model and the spatial-domain SISR baselines.}
  \label{tab:training}
\end{table*}

\vspace{0.3cm}

\begin{table*}[tb]
  \centering
  \footnotesize
  \begin{tabular}{lcccc}
    \toprule
    Method & Inference preprocessing time (ms) & Model inference time (ms) & Avg. PSNR & Avg. SSIM \\
    \midrule
    EDSR RGB & 1.9 & 0.7 & 35.11 & 0.9717 \\
    EDSR Y & 2.2 & 0.8 & 35.81 & 0.9749 \\
    \textbf{FreqSR (ours)} & 2.3 & 0.7 & 29.35 & 0.7213 \\
    \bottomrule
  \end{tabular}
  \caption{Inference performance comparison between our proposed SISR pipeline based on FreqSR model and spatial-domain SISR baselines.}
  \label{tab:inference}
\end{table*}

\begin{figure*}[!ht]
  \centering
  \begin{minipage}[b]{0.3\linewidth}
    \centering
    \includegraphics[width=\linewidth]{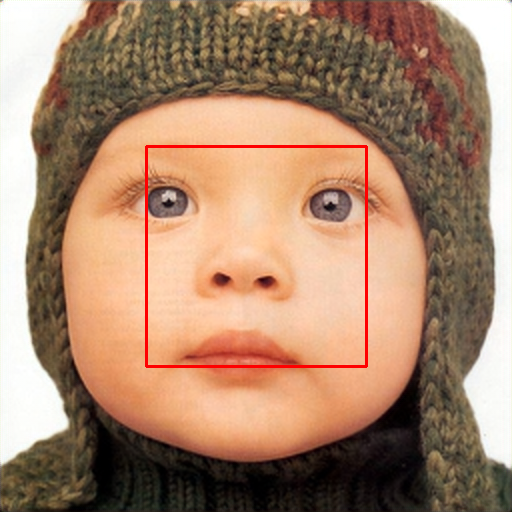}
    \centerline{(a) EDSR RGB, PSNR: 38.32 dB}
  \end{minipage}
  \hfill
  \begin{minipage}[b]{0.3\linewidth}
    \centering
    \includegraphics[width=\linewidth]{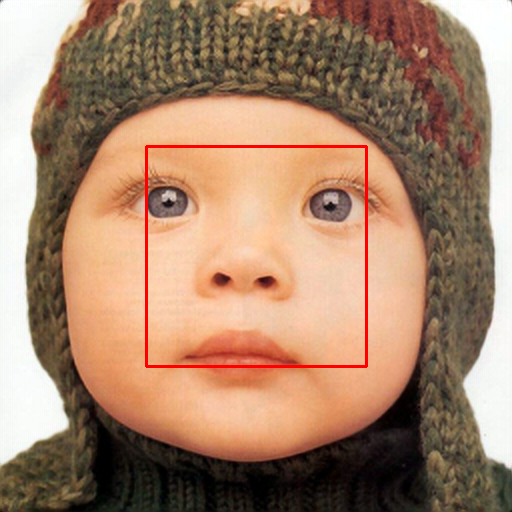}
    \centerline{(b) EDSR Y, PSNR: 38.62 dB}
  \end{minipage}
  \hfill
  \begin{minipage}[b]{0.3\linewidth}
    \centering
    \includegraphics[width=\linewidth]{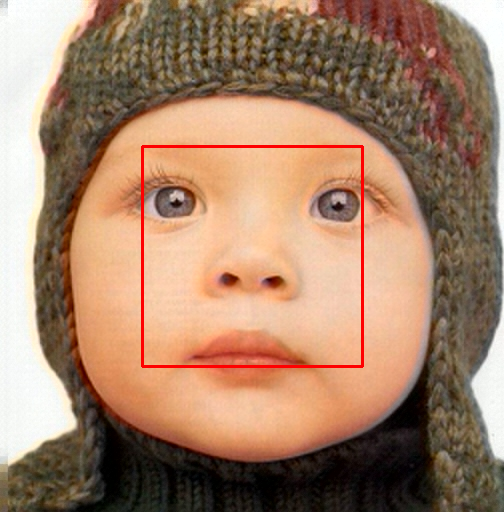}
    \centerline{(c) FreqSR (ours), PSNR: 30.85 dB}
  \end{minipage}

  \vspace{0.3cm}

  \begin{minipage}[b]{0.3\linewidth}
    \centering
    \includegraphics[width=\linewidth]{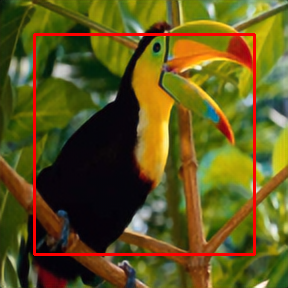}
    \centerline{(d) EDSR RGB, PSNR: 34.70 dB}
  \end{minipage}
  \hfill
  \begin{minipage}[b]{0.3\linewidth}
    \centering
    \includegraphics[width=\linewidth]{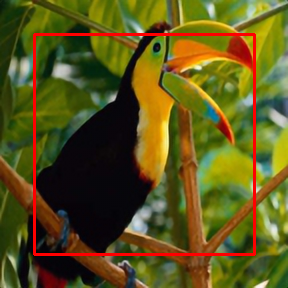}
    \centerline{(e) EDSR Y, PSNR: 36.34 dB}
  \end{minipage}
  \hfill
  \begin{minipage}[b]{0.3\linewidth}
    \centering
    \includegraphics[width=\linewidth]{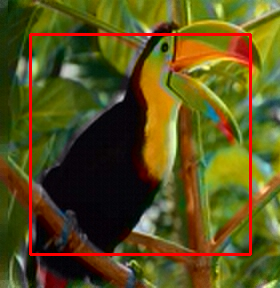}
    \centerline{(f) FreqSR (ours), PSNR: 29.24 dB}
  \end{minipage}

  \caption{Comparison of HR images generated from the Set5 dataset. (a)-(c) and (d)-(f) show two example model outputs from EDSR RGB, EDSR Y, and our proposed SISR pipeline based on FreqSR model, respectively. PSNR values are computed on 220×220 center crops of the HR outputs. The inference results of FreqSR model are visually similar to those of other methods.}
  \label{fig:qualitative}
\end{figure*}

\subsection{Inference Pipeline}
Table \ref{tab:inference} summarizes the performance of our proposed frequency domain SISR pipeline using FreqSR model during inference, comparing it with the spatial-domain SISR baseline approaches EDSR RGB and EDSR Y. Computational efficiency is assessed by measuring preprocessing time and model inference time, while image quality is evaluated using average peak signal-to-noise ratio (PSNR) and average structural similarity index measure (SSIM) on the 220x220 center crop of Set5 dataset. Fig.\ref{fig:qualitative} presents two examples from the Set5 dataset, comparing HR outputs from the spatial domain SISR baselines and our proposed pipeline.

The frequency-based SISR pipeline introduces slight preprocessing overhead due to DCT coefficient upsampling. However, FreqSR inference latency remains comparable to spatial-domain SISR baseline approaches despite this additional preprocessing. Overall inference pipeline timing is challenging to measure precisely due to required CPU-based post-processing operations (e.g., DCT-to-RGB conversion). Nevertheless, FreqSR maintains the memory efficiency advantages of compressed JPEG DCT representations. Although inference latency improvements are less pronounced compared to training gains, these results highlight the potential of frequency-domain methods in scenarios where data loading efficiency and memory optimization are critical.

In terms of image quality, the average PSNR and SSIM scores for FreqSR model are lower than the EDSR RGB and EDSR Y baselines. This is expected, as reconstructing high-frequency details directly from JPEG DCT coefficients is inherently challenging. Besides, minor misalignment between the luminance and chrominance channels can introduce subtle artifacts, which may impact SSIM -- a pixel-level metric that is sensitive to such discrepancies. These misalignments are often the result of block rounding inherent to JPEG’s 8×8 DCT block structure. Current padding methods provide limited correction. Enhancing the reconstruction pipeline to mitigate these effects presents promising future work.

\section{Conclusion \& Future Directions}
\label{sec:conclusion}
We presented a lightweight frequency-domain SISR pipeline that operates directly on JPEG DCT coefficients, bypassing full JPEG decoding and RGB reconstruction. The framework incorporates FreqSR, a new model architecture designed for frequency-domain processing. Our proposed SISR pipeline enables faster data loading (2.6×) and training (2.5×) while maintaining competitive reconstruction quality. By leveraging intermediate representations, our approach reduces computational overhead and memory usage -- key challenges in edge and low-power AI deployments. Qualitative results demonstrate consistent visual fidelity with acceptable perceptual trade-offs, making our proposed SISR pipeline using FreqSR architecture suitable for real-time and on-device applications where efficiency is paramount. 

We hope this early study contributes to the growing body of research on compressed-domain learning and highlights the promise of computation-aware design for AI on resource-constrained devices. Future directions include extending the method to video super-resolution, exploring compact frequency-domain architectures optimized for low-power and real-time deployment, and assessing the feasibility of applying the proposed approach to image codecs beyond JPEG. Additionally, we aim to investigate the integration of our pipeline with existing edge AI frameworks to further enhance its applicability. We also plan to conduct more analysis to validate the robustness and efficiency of our approach.





\newpage
{
\setlength{\itemsep}{0pt} 
\setlength{\parskip}{0pt} 
\bibliographystyle{IEEEbib}
\bibliography{strings,refs}

@article{howard2017mobilenets,
  title={Mobilenets: Efficient convolutional neural networks for mobile vision applications},
  author={Howard, Andrew G},
  journal={arXiv preprint arXiv:1704.04861},
  year={2017}
}

@article{iandola2016squeezenet,
  title={SqueezeNet: AlexNet-level accuracy with 50x fewer parameters and< 0.5 MB model size},
  author={Iandola, Forrest N},
  journal={arXiv preprint arXiv:1602.07360},
  year={2016}
}

@inproceedings{jacob2018quantization,
  title={Quantization and training of neural networks for efficient integer-arithmetic-only inference},
  author={Jacob, Benoit and others},
  booktitle={IEEE conference on computer vision and pattern recognition},
  pages={2704--2713},
  year={2018}
}

@article{denton2014exploiting,
  title={Exploiting linear structure within convolutional networks for efficient evaluation},
  author={Denton, Emily L and others},
  journal={Advances in neural information processing systems},
  volume={27},
  year={2014}
}

@misc{nvidia_dali,
  author       = {NVIDIA},
  title        = {DALI: The NVIDIA Data Loading Library},
  year         = {2025},
  howpublished = {\url{https://github.com/NVIDIA/DALI}},
  note         = {Accessed: 2024-06-03}
}

@misc{rocAL,
  author       = {AMD},
  title        = {{rocAL}: {ROCm} Augmentation Library},
  year         = {2025},
  howpublished = {\url{https://github.com/ROCmSoftwarePlatform/rocAL}},
  note         = {Accessed: 2024-06-03}
}

@article{gueguen2018faster,
  title={Faster neural networks straight from jpeg},
  author={Gueguen, Lionel and others},
  journal={Advances in Neural Information Processing Systems},
  volume={31},
  year={2018}
}

@inproceedings{xu2020learning,
  title={Learning in the frequency domain},
  author={Xu, Kai and others},
  booktitle={Conference on computer vision and pattern recognition},
  pages={1740--1749},
  year={2020}
}

@inproceedings{deguerre2019fast,
  title={Fast object detection in compressed jpeg images},
  author={Deguerre, Benjamin and others},
  booktitle={2019 ieee intelligent transportation systems conference (itsc)},
  pages={333--338},
  year={2019},
  organization={IEEE}
}

@inproceedings{park2023rgb,
  title={{RGB} no more: Minimally-decoded {JPEG} Vision Transformers},
  author={Park, Jeongsoo and Johnson, Justin},
  booktitle={Conference on Computer Vision and Pattern Recognition},
  pages={22334--22346},
  year={2023}
}

@inproceedings{lim2017enhanced,
  title={Enhanced deep residual networks for single image super-resolution},
  author={Lim, Bee and others},
  booktitle={IEEE conference on computer vision and pattern recognition workshops},
  pages={136--144},
  year={2017}
}

@inproceedings{ledig2017photo,
  title={Photo-realistic single image super-resolution using a generative adversarial network},
  author={Ledig, Christian and others},
  booktitle={IEEE conference on computer vision and pattern recognition},
  pages={4681--4690},
  year={2017}
}

@inproceedings{zhang2021edge,
  title={Edge-oriented convolution block for real-time super resolution on mobile devices},
  author={Zhang, Xindong and others},
  booktitle={ACM International Conference on Multimedia},
  pages={4034--4043},
  year={2021}
}

@inproceedings{liang2021swinir,
  title={Swinir: Image restoration using swin transformer},
  author={Liang, Jingyun and Cao, Jiezhang and Sun, Guolei and Zhang, Kai and Van Gool, Luc and Timofte, Radu},
  booktitle={Proceedings of the IEEE/CVF international conference on computer vision},
  pages={1833--1844},
  year={2021}
}

@article{wallace1991jpeg,
  title={The JPEG still picture compression standard},
  author={Wallace, Gregory K},
  journal={Communications of the ACM},
  volume={34},
  number={4},
  pages={30--44},
  year={1991},
  publisher={AcM New York, NY, USA}
}

@article{cai2021freqnet,
  title={FreqNet: A Frequency-domain Image Super-Resolution Network with Dicrete Cosine Transform},
  author={Cai, Runyuan and others},
  journal={arXiv preprint arXiv:2111.10800},
  year={2021}
}

@inproceedings{timofte2017ntire,
  title={Ntire 2017 challenge on single image super-resolution: Methods and results},
  author={Timofte, Radu and others},
  booktitle={IEEE conference on computer vision and pattern recognition workshops},
  pages={114--125},
  year={2017}
}

@inproceedings{bevilacqua2012low,
  TITLE = {{Low-Complexity Single-Image Super-Resolution based on Nonnegative Neighbor Embedding}},
  AUTHOR = {Bevilacqua, Marco and others},
  BOOKTITLE = {{British Machine Vision Conference (BMVC)}},
  YEAR = {2012},
  MONTH = Sep,
}

@misc{libjpeg,
  author       = {Independent JPEG Group},
  title        = {Libjpeg},
  year         = {2024},
  howpublished = {\url{https://www.ijg.org/}},
  note         = {Accessed: 2025-06-03}
}

@inproceedings{ehrlich2020quantization,
  title={Quantization guided jpeg artifact correction},
  author={Ehrlich, Max and Davis, Larry and Lim, Ser-Nam and Shrivastava, Abhinav},
  booktitle={ECCV},
  pages={293--309},
  year={2020},
  organization={Springer}
}

@inproceedings{lu2022transformer,
  title={Transformer for single image super-resolution},
  author={Lu, Zhisheng and others},
  booktitle={Conference on computer vision and pattern recognition},
  pages={457--466},
  year={2022}
}

@inproceedings{lo2019exploring,
  title={Exploring semantic segmentation on the dct representation},
  author={Lo, Shao-Yuan and Hang, Hsueh-Ming},
  booktitle={Conference on Multimedia in Asia},
  pages={1--6},
  year={2019}
}

@article{zolnouri2020importance,
  author = {Zolnouri, Mahdi and others},
  pages = {1-10},
  title = {Importance of data loading pipeline in training deep neural networks},
  journal      = {CoRR},
  year = {2020},
}

@inproceedings{yang2019accelerating,
  title={Accelerating data loading in deep neural network training},
  author={Yang, Chih-Chieh and Cong, Guojing},
  booktitle={2019 HiPC},
  pages={235--245},
  year={2019},
}

@inproceedings{ehrlich2019deep,
  title={Deep residual learning in the jpeg transform domain},
  author={Ehrlich, Max and Davis, Larry S},
  booktitle={International conference on computer vision},
  pages={3484--3493},
  year={2019}
}

@inproceedings{chen2022mobile,
  title={Mobile-former: Bridging mobilenet and transformer},
  author={Chen, Yinpeng and Dai, Xiyang and Chen, Dongdong and Liu, Mengchen and Dong, Xiaoyi and Yuan, Lu and Liu, Zicheng},
  booktitle={Proceedings of the IEEE/CVF conference on computer vision and pattern recognition},
  pages={5270--5279},
  year={2022}
}
}
\end{document}